\documentclass[runningheads]{llncs}
\usepackage[T1]{fontenc}
\usepackage{bbding}
\usepackage{booktabs}
\usepackage{multirow}
\usepackage{graphicx}
\usepackage{makecell}
\usepackage{amsfonts}
\usepackage{amsmath}
\usepackage{hyperref}

\usepackage{graphicx,verbatim}
\begin{document}
\title{Anatomy-Guided Residual Motion Diffusion for Controllable 4D Cardiac MRI Synthesis}

\titlerunning{Controllable 4D Cardiac MRI Synthesis}
%
\author{Yiheng Cao \inst{1}\orcidID{0009-0001-5271-0894} \and
Gustavo Andrade-Miranda\inst{2}\orcidID{0000-0002-6499-5655} \and
Jiatian Zhang \inst{1} \and
Lingxiao Zhao \inst{3} \and
Xin Gao\inst{1}}
\index{Cao, Yiheng}
\index{Gustavo, Andrade-Miranda}
\index{Zhang, Jiatian}
\index{Zhao, Lingxiao}
\index{Gao, Xin}

\authorrunning{Y. Cao et al.}
%
\institute{Suzhou Institute of Biomedical Engineering and Technology, Chinese Academy of Science, Suzhou, China\\
\email{caoyh@sibet.ac.cn} \and
Systemes Complexes et Intelligence Artificielle, IMT Mines Ales, Ales, France \and
School of Artificial Intelligence and Advanced Computing, Xi'an Jiaotong-Liverpool University, Suzhou, China }
  
\maketitle              
\begin{abstract}
Developing robust artificial intelligence models for 4D (3D + time) medical imaging is constrained by limited annotated data, inter-device domain shifts, and privacy restrictions. To address this, we propose a 4D controllable generative framework for anatomically consistent data augmentation. A semi-supervised variational autoencoder learns a compact latent representation of anatomical volumes while jointly predicting aligned segmentation masks in a unified framework. Anatomical structure is then disentangled from temporal dynamics through a cascaded latent diffusion model (LDM). A static LDM generates subject-specific anatomy conditioned on clinical priors (diagnosis and volumes measures) and a subsequent motion LDM estimates residual latent motions, ensuring strict temporal coherence across the 4D sequence.
The proposed approach was evaluated on cine cardiac MRI as a representative 4D imaging application. Experiments across multiple datasets demonstrate high controllability of static anatomy (Pearson $r > 0.8$) and strong temporal coherence (FVD = $288.08$). In cross-vendor generalization experiments, augmenting training sets with synthetic 4D sequences significantly improves downstream segmentation performance. Using nnU-Net, the proposed augmentation strategy improves the average Dice score by 1.4\% and reduces the Hausdorff Distance by 3.0~mm compared to training on real data alone, for the left ventricle, Dice improves by 2.8\% with a 5.4~mm reduction in boundary error.
Overall, this framework provides a scalable and controllable solution for 4D medical image synthesis, supporting the development of more robust models with limited annotations and cross-vendor variability. Code available on \href{GitHub}{https://github.com/cyiheng/4DCardiacMRISynthesis}.

\keywords{4D cardiac MRI  \and Generation model \and Latent diffusion model}

\end{abstract}
\section{Introduction}

Cardiovascular diseases remain the leading cause of mortality worldwide and require accurate diagnosis. Cine cardiac magnetic resonance (CMR) imaging is the gold standard for assessing cardiac function, providing high-resolution spatiotemporal data on ventricular volumes and myocardial wall motion. Recent advances in deep learning (DL) have improved automated CMR analysis \cite{bernard_deep_2018}, data acquisition and sharing remain limited by privacy regulations, resulting in small multi-center datasets and domain shifts that impair generalization \cite{guo_impact_2024,kaissis_secure_2020}. DL methods also rely on large, expertly annotated datasets, a constraint that is particularly critical for 4D (3D + time) data, where full-cycle delineation is labor-intensive and time-consuming \cite{tajbakhsh_embracing_2020}. Consequently, annotations are often restricted to end-diastolic (ED) and end-systolic (ES) phases, discarding substantial spatiotemporal information \cite{bernard_deep_2018,tajbakhsh_embracing_2020}. 

Generative modeling offers a potential solution through realistic data synthesis \cite{gao_synthetic_2023,koetzier_generating_2024}. Medical image generation relies on both physics-based simulations and data-driven approaches, with modern deep generative models learning high-dimensional statistical distributions directly from data \cite{frangi_simulation_2018,kazerouni_diffusion_2023,wang_generative_nodate}. However, 4D cine CMR generation remains challenging due to high dimensionality and the need for physiologically plausible cardiac motion. Early GAN-based methods either model cine dynamics via latent trajectories with limited anatomical control or rely on label-conditioned synthesis that does not scale to volumetric 4D data without dense supervision \cite{amirrajab_label-informed_2022,vukadinovic_gancmri_2023}. More recent diffusion-based approaches improve perceptual fidelity and enable conditional generation using clinical attributes or textual prompts \cite{liu_texdc_2025,you_temporal_2026}. However, most existing methods combine anatomical structures and temporal dynamics in a single generative process, limiting independent controllability and long-term consistency. Additionally, many approaches either focus solely on intensity generation without paired segmentation masks or depend on densely annotated, often proprietary datasets, limiting scalability, reproducibility, and cross-institutional validation \cite{dou_4d_2026,you_temporal_2026}. Despite promising visual realism, current generative models remain insufficient for producing anatomically controllable, label-aware 4D cine CMR data suitable for robust downstream learning tasks.

In this work, we bridge these gaps by proposing a robust 4D generation pipeline for cine CMR. We simplify the 4D synthesis problem by decoupling it into static anatomical generation and residual latent motion prediction. Our contribution is threefold:
\begin{itemize}
\item We introduce a semi-supervised variational autoencoder (VAE) framework that jointly models anatomical structure and semantic information from partially labeled datasets, enabling the direct generation of paired intensity volumes and segmentation masks without dense manual annotations for all samples.

\item We propose a residual latent motion model that characterizes cardiac dynamics as residual trajectories within the latent space. The model decouples static anatomy from cardiac dynamics, generating temporally coherent 4D cine sequences while preserving anatomical consistency.

\item We demonstrate that adding synthetic data improves the cross-vendor generalization of cardiac segmentation models by reducing left ventricle (LV), right ventricle (RV), and myocardium (MYO) boundary errors, enabling reliable clinical quantification of ventricular volumes and ejection fraction.


\end{itemize}

\begin{figure}[ht]
    \centering
    \includegraphics[width=\linewidth]{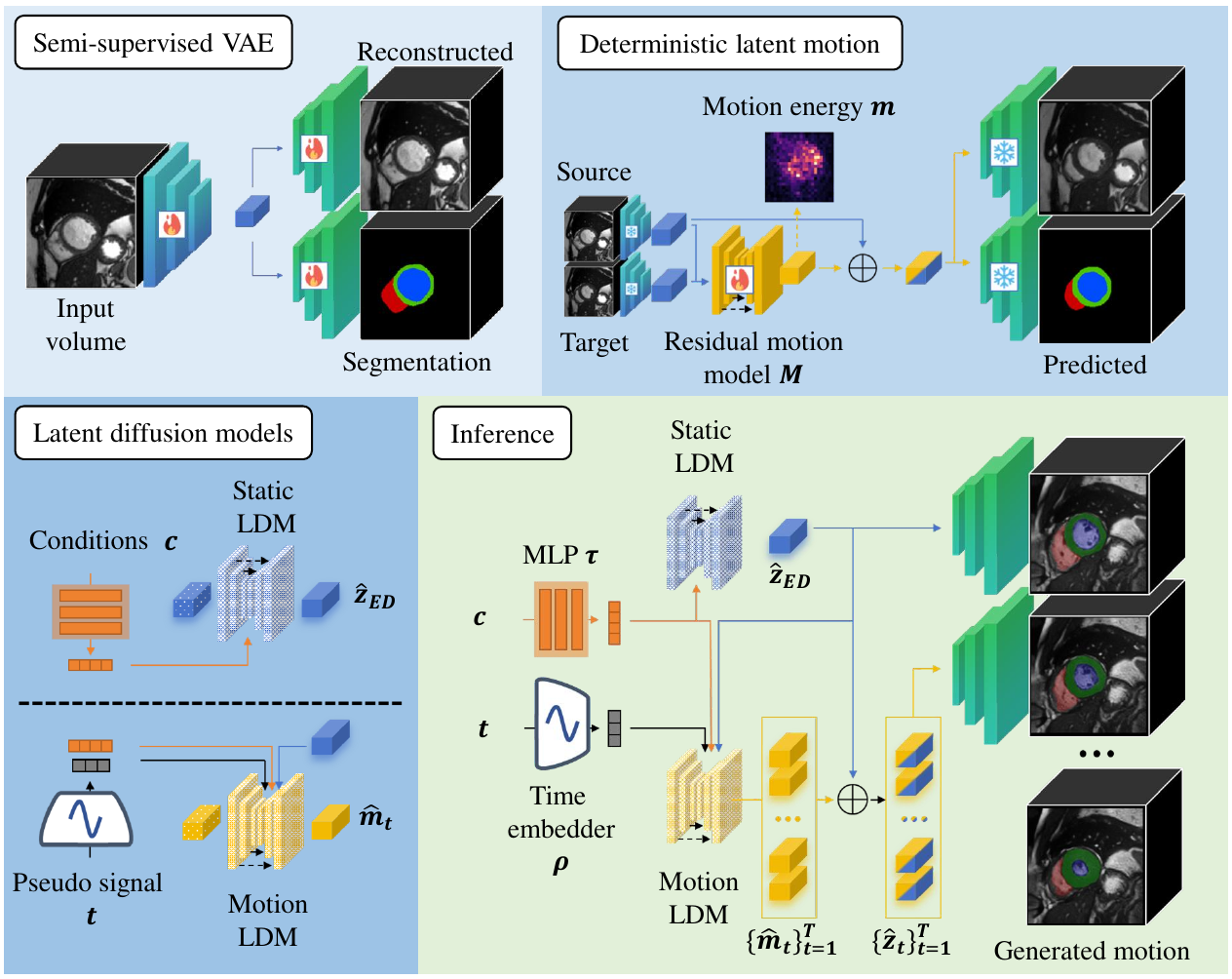}
    \caption{Overview of the proposed framework. Blue panels: VAE compression, deterministic residual motion extraction, and cascaded LDM training. Green panel: the static LDM generates base anatomy $\hat{z}_{ED}$ from conditions $\mathbf{c}$, while the motion LDM predicts temporal residuals $\hat{m}_t$. These are aggregated ($\hat{z}_{ED} + \hat{m}_t$) and passed through the VAE decoders to produce the final 4D volumes and inherently aligned segmentation masks.}
    \label{fig:overview}
\end{figure}

\section{Methods}

Let $\mathbf{x} \in \mathbb{R}^{H \times W \times D \times T}$ denote a 4D cardiac MRI sequence with $T$ frames. We model $p(\mathbf{x} | \mathbf{c})$ conditioned on clinical priors $\mathbf{c}$ (e.g., diagnosis, volumetric indices) through three components: (1) a semi-supervised VAE for latent compression, (2) a static latent diffusion model (LDM) for anatomical generation, and (3) a residual motion LDM for temporal dynamics (Figure \ref{fig:overview}).

\subsection{Semi-supervised Latent Representation}

The input space is first map to a low-dimensional latent space $\mathcal{Z}$. We use a 3D VAE-GAN with a shared encoder $E$ and two decoders: an intensity reconstruction decoder $D_{rec}$ and a segmentation decoder $D_{seg}$. Given an input volume $x$, the encoder produces a latent distribution $z~\sim~E(x)$. The training objective is a composite loss that ensures reconstruction fidelity, latent regularization, semantic alignment, and textural realism:

\begin{equation}
    \begin{split}
        \mathcal{L}_{VAE} &= \lambda_{pixel}\mathcal{L}_{pixel}(x, D_{rec}(z)) + \lambda_{KL}\mathcal{L}_{KL}(q(z|x)||p(z)) \\
        &\quad + \mathbb{I}_{label} \cdot \lambda_{seg}\mathcal{L}_{seg}(y, D_{seg}(z)),
    \end{split}
\end{equation}
where $\mathcal{L}_{pixel}$ includes $L_1$, perceptual, and adversarial terms, $\mathcal{L}_{KL}$ regularizes the latent posterior $q(z|x)$ towards a standard Gaussian prior $p(z)$ to facilitate valid sampling. 
The cross-entropy $\mathcal{L}_{seg}$ is applied only when labeled data $y$ are available ($\mathbb{I}_{label} = 1$), allowing the model to leverage both labeled and unlabeled datasets. 

\subsection{Static Anatomy Generation}

We train a static LDM generates the ED latent base anatomy $\hat{z}_{ED}$ conditioned on clinical priors $\mathbf{c}$ (categorical diagnosis, slice count, and ED/ES/ejection fraction volumes). A Multi-Layer Perceptron (MLP) $\tau$ projects $\mathbf{c}$ to condition the LDM through cross-attention. To handle variable slice counts within a fixed latent resolution, a through-plane binary mask indicating valid slices is concatenated during training and sampling. This prevents the model from learning padded regions, ensuring center-cropped decoded volumes remain anatomically consistent without boundary artifacts.

\subsection{Residual Motion Dynamics Generation}

We model the temporal evolution of the cardiac cycle as a residual offset relative to the ED frame. For any time frame $t$, we suppose that the latent representation $z_t$ can be expressed as $z_t = z_{ED} + m_t$, where $m_t$ represents the residual latent motion. Motion modeling is performed in two stages: deterministic residual extraction and stochastic residual generation.

\subsubsection{Learning the Residual Latent Motion Space.}
A deterministic latent motion predictor $M$ is trained to learn the residual mapping between paired frames $(z_{ED},z_t)$. The network predicts: 
$\hat{m}_t = M(z_{ED},z_t)$ with $\hat{z}_{t} = z_{ED} + \hat{m}_t$.
To preserve high-frequency details and semantic integrity, the model is supervised in both latent and image space using:
\begin{equation}
\mathcal{L}_{motion} = \lambda_{pixel}\mathcal{L}_{pixel}(x_{t}, D_{rec}(\hat{z}_{t})) + \mathbb{I}_{label} \cdot \lambda_{seg}\mathcal{L}_{seg} + \mathcal{L}_{reg}
\end{equation}
The regularization term
\begin{equation}
    \mathcal{L}_{reg} = \lambda_{sparse} | \hat{m} |^2_2 + \lambda_{identity} | M(z_{ED}, z_{ED}) |_1
\end{equation}
encourages minimal latent motion energy and enforces an identity constraint. This ensures zero residuals for identical input frames and prevents autoencoding behavior, forcing $M$ to learn true relative motion to preserve anatomical consistency.

\subsubsection{Motion Diffusion Model.}
Once trained, the frozen $M$ extracts residuals $m_t$ to train a conditional motion LDM. This LDM is conditioned on (i) the static reference $z_{ED}$ through concatenation, (ii) the clinical embeddings $\tau(\mathbf{c})$, and (iii) a sinusoidal time embeddings $\rho(t)$ representing the normalized cardiac phase. 

During inference, the pipeline samples $\hat{z}_{ED}$ from the static LDM, then samples a sequence of residuals $\{\hat{m}_t\}^T_{t=1}$ from the motion LDM. These residuals are aggregated and passed through the pre-trained VAE decoders to jointly reconstruct the final 4D intensity sequence and its inherently aligned segmentation masks.

\section{Experiments and Results}

\subsubsection{Datasets.}
Our framework was trained on a combined dataset of 956 patients, including 100 patients with annotations at ED and ES from the ACDC dataset \cite{bernard_deep_2018} and 856 unlabeled patients from the Kaggle Data Science Bowl (DSB) dataset \cite{noauthor_second_nodate}. The semi-supervised VAE was trained using all cases for reconstruction, while segmentation supervision was provided exclusively by the 100 labeled ACDC cases. Internal evaluation was performed on 50 patients of the ACDC test set and a held-out subset of 50 patients from DSB dataset. Cross-vendor generalization was assessed using an external test sets: 134 and 157 patients from the M\&Ms and M\&Ms2, respectively \cite{campello_multi-centre_2021,martin-isla_deep_2023}.

\subsubsection{Implementation details.} 
All models were implemented in PyTorch using a single NVIDIA RTX 4090 GPU. The images were resampled at a fixed spacing of $1 \times 1 \times 10$ mm and centrally cropped to $192 \times 192 \times Z$ where $Z \in [5,16]$ denotes the number of slices. Intensities were clipped to the $[0.5, 99.5]$ percentiles and normalized to the range $[-1, 1]$. To handle variable slice counts in $Z$, volumes are padded using symmetric edge replication. This preserves boundary intensities, avoiding artificial artifacts and bias in the latent space, while ensuring anatomically consistent borders without introducing unintended intensity changes during training.

The semi-supervised VAE was trained for 100{,}000 steps using AdamW ($lr = 10^{-4}$), with $\lambda_{seg} = 20.0$ and $\lambda_{KL} = 10^{-7}$. The reconstruction loss combined $L_1$, perceptual, and adversarial terms weighted at 1.0, 0.3, and 0.1, respectively. The latent motion predictor was trained for 100{,}000 steps (AdamW, $lr = 2 \times 10^{-4}$) with $\lambda_{seg} = 10.0$, $\lambda_{sparse} = 0.01$, $\lambda_{identity} = 1.0$, and a reconstruction loss weighted at 2.0, 1.0, and 0.01 ($L_1$, perceptual, adversarial).
Both LDM were trained for 200{,}000 steps using an 8-bit AdamW optimizer ($lr = 2 \times 10^{-5}$) and DDIM sampling with 50 inference steps. Static LDM training included random affine scaling and pseudo-ED sampling, with clinical volumes dynamically recomputed using frozen VAE segmentation outputs.

\subsubsection{Evaluation metrics.}

We evaluate the framework based on perceptual quality, anatomical controllability, and utility for downstream segmentation task. 
Perceptual fidelity and temporal coherence are quantified using the Fréchet Inception Distance (FID) and Fréchet Video Distance (FVD), respectively. FID is computed on static ED volumes, while FVD is evaluated on full cine sequences. FID is computed using 10-fold bootstrapping with 2{,}000 generated volumes and 956 real training cases. For FVD, 1,000 synthetic 4D sequences were unpacked along the longitudinal (z) axis into temporal 2D+t video clips and compared against real training clips. FVD uncertainty was quantified via 100-iteration non-parametric bootstrap resampling at the video level.
Anatomical controllability was evaluated by calculating the Pearson correlation between the clinical priors input and synthetic measurements across 20 cases per pathology using three classifier-free guidance (CFG) scales. 
To assess practical clinical utility, we use semantic segmentation as a representative downstream task to rigorously validate that synthetic data augmentation preserves anatomical boundaries and enhances cross-domain robustness. Segmentation models (nnU-Net \cite{isensee_nnu-net_2021}, UNETR \cite{hatamizadeh_unetr_2022}, and Swin-UNETR \cite{hatamizadeh_swin_2022}) are trained using 900 synthetic ED/ES volume pairs and evaluated using Dice Similarity Coefficient (DSC) and 95\% Hausdorff Distance (HD), reported as mean and standard deviation over five runs on internal (ACDC) and external (M\&Ms, M\&Ms2) test sets.

\subsubsection{Generation fidelity.}

\begin{figure}[t]
    \centering
    \includegraphics[width=\linewidth]{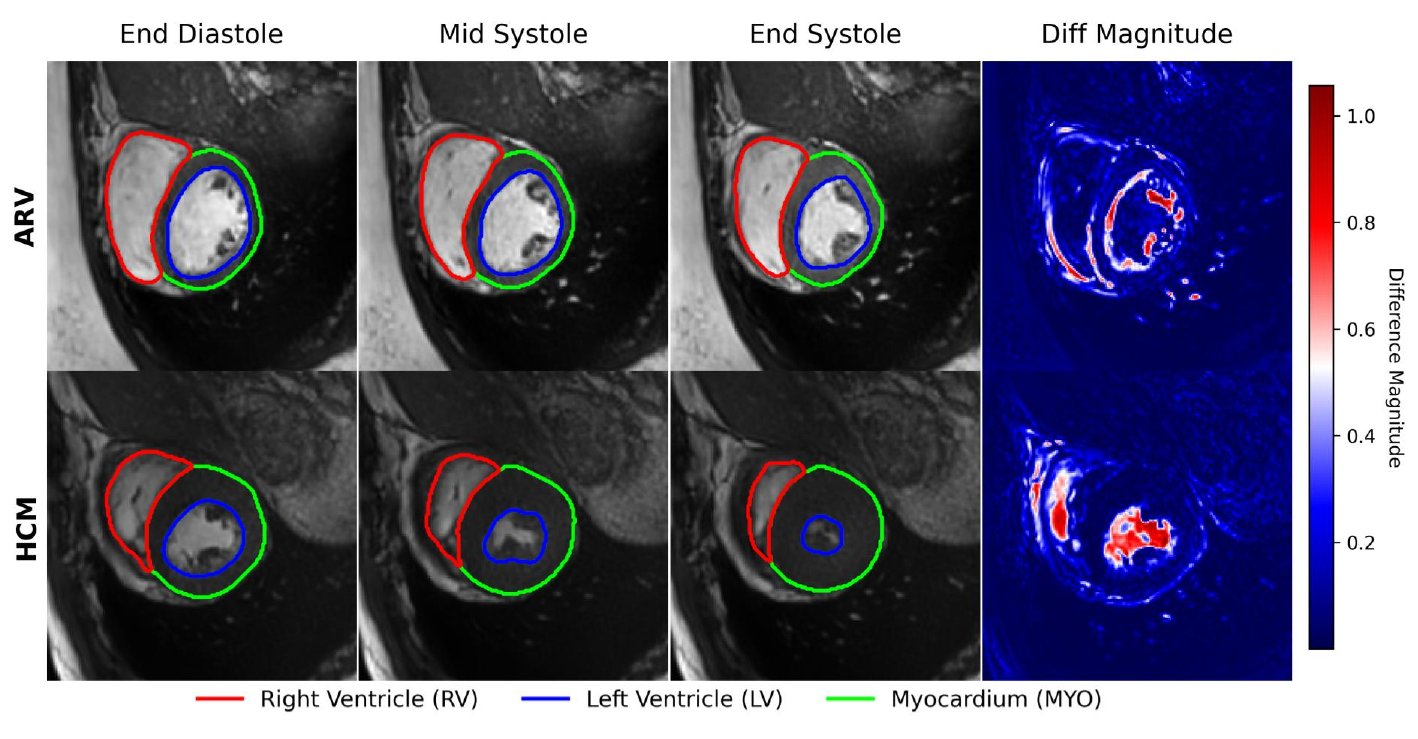}
    \caption{Example of synthetic volumes and segmentation across pathological classes.}
    \label{fig:visual}
\end{figure}

The Figure \ref{fig:visual} displays synthetic sequences for abnormal right ventricle (ARV) and hypertrophic cardiomyopathy (HCM) pathologies (full motion available in the Supplementary Material). The framework reproduces anatomical features, such as myocardial thickening in HCM, with segmentation masks that remain aligned with generated intensities across the cardiac cycle. The difference magnitude maps demonstrate substantial regional changes. Quantitative perceptual metrics support visual assessment, with a slice-wise FID of $72.21 \pm 0.95$ and FVD of $288.08 \pm 3.63$. These results suggest that the generated sequences exhibit realistic intensity distributions and stable temporal evolution.
\begin{table}[hb]
\caption{Segmentation performance using nnU-Net as backbone.}
\label{tab:nnunet_seg}
\resizebox{\textwidth}{!}{%
\begin{tabular}{clcccccc}
\hline
\multirow{2}{*}{\textbf{Dataset}} & \multicolumn{1}{c}{\multirow{2}{*}{\textbf{Data}}} & \multicolumn{1}{l}{\textbf{}} & \textbf{DSC}          & \multicolumn{1}{l}{\textbf{}} & \multicolumn{1}{l}{}   & \textbf{HD}            & \multicolumn{1}{l}{}   \\ \cline{3-8} 
                                  & \multicolumn{1}{c}{}                               & \textbf{RV}                   & \textbf{MYO}          & \textbf{LV}                   & \textbf{RV}            & \textbf{MYO}           & \textbf{LV}            \\ \hline
                                  & nnU-Net                                           & \textbf{91.02 ± 0.08}         & \textbf{88.90 ± 0.04} & 93.18 ± 0.17                  & 4.95   ± 0.13          & 2.87   ± 0.09          & 3.37   ± 0.33          \\
ACDC                              & +Synth ED                                         & 90.84 ± 0.09                  & 88.34 ± 0.02          & 93.23 ± 0.05                  & 4.15   ± 0.34          & 2.78   ± 0.04          & 3.00   ± 0.06          \\
                                  & +Synth ED/ES                                    & 90.95 ± 0.11                  & 88.74 ± 0.05          & \textbf{93.90 ± 0.06}         & \textbf{4.12   ± 0.29} & \textbf{2.70   ± 0.06} & \textbf{2.77   ± 0.09} \\ \hline
                                  & nnU-Net                                           & 87.08 ± 0.24                  & 81.83 ± 0.14          & 84.87 ± 0.78                  & 7.64   ± 0.47          & 8.60   ± 0.36          & 14.57   ± 0.97         \\
M\&Ms                             & +Synth ED                                         & \textbf{87.55   ± 0.07}       & 82.34 ± 0.10          & 87.54 ± 0.30                  & 6.62   ± 0.14          & 6.59   ± 0.16          & 8.99   ± 0.50          \\
                                  & +Synth ED/ES                                    & 87.44   ± 0.18                & \textbf{82.52 ± 0.10} & \textbf{88.27 ± 0.18}         & \textbf{6.55   ± 0.28} & \textbf{6.49   ± 0.14} & \textbf{8.45   ± 0.36} \\ \hline
                                  & nnU-Net                                           & 85.93 ± 0.27                  & 80.00 ± 0.15          & 86.21 ± 0.49                  & 10.44   ± 0.81         & 9.00   ± 0.22          & 13.71   ± 1.10         \\
M\&Ms2                            & +Synth ED                                         & 86.33 ± 0.16                  & 80.57 ± 0.15          & 88.09 ± 0.12                  & 9.14   ± 0.11          & 6.78   ± 0.12          & 9.44   ± 0.25          \\
                                  & +Synth ED/ES                                    & \textbf{86.60 ± 0.42}         & \textbf{81.12 ± 0.78} & \textbf{88.35 ± 0.19}         & \textbf{9.12   ± 1.29} & \textbf{6.77   ± 0.25} & \textbf{8.98   ± 0.41} \\ \hline
\end{tabular}%
}
\end{table}

\begin{figure}[t]
    \centering
    \includegraphics[width=\linewidth]{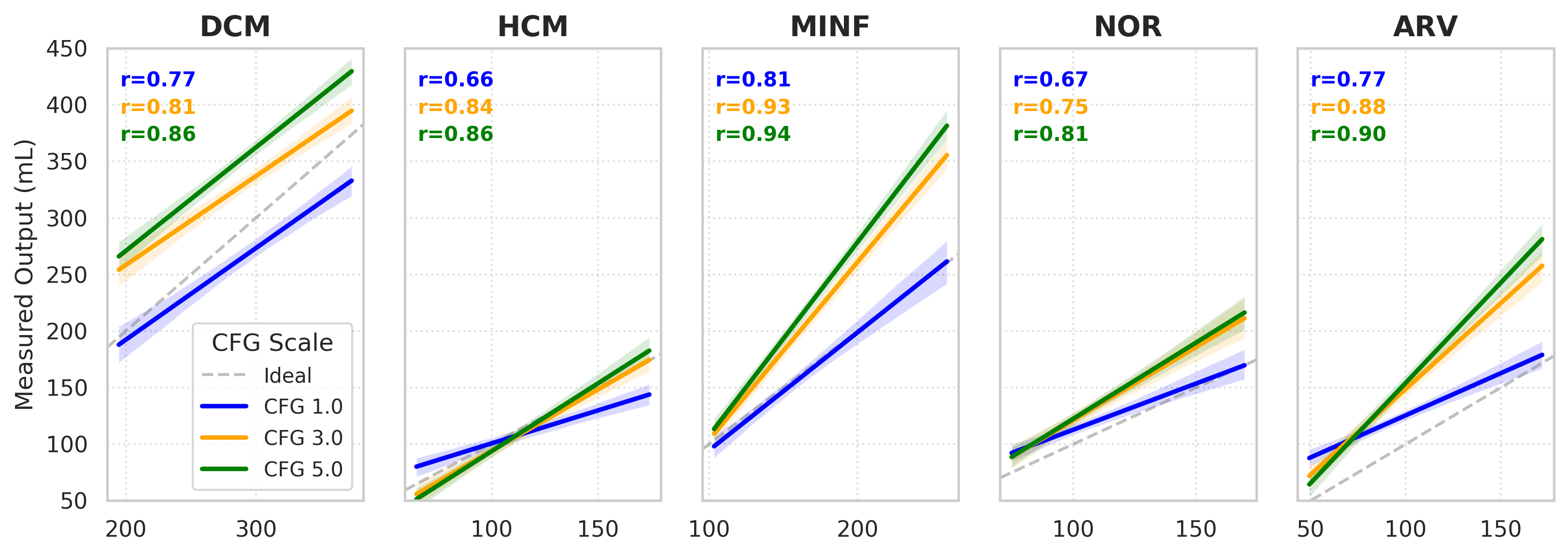}
    \caption{Correlation between input clinical volume priors and measured synthetic volumes across different pathologies and CFG scales.}
    \label{fig:corr}
\end{figure}
\begin{table}[ht]
\caption{Segmentation performance using UNETR and Swin-UNETR as backbone.}
\label{tab:altseg}
\resizebox{\textwidth}{!}{%
\begin{tabular}{@{}clcccccc@{}}
\toprule
\multirow{2}{*}{\textbf{Dataset}}           & \multicolumn{1}{c}{\multirow{2}{*}{\textbf{Data}}} & \multicolumn{1}{l}{\textbf{}} & \textbf{DSC}          & \multicolumn{1}{l}{\textbf{}} & \multicolumn{1}{l}{}  & \textbf{HD}           & \multicolumn{1}{l}{}  \\ \cmidrule(l){3-8} 
                                            & \multicolumn{1}{c}{}                               & \textbf{RV}                   & \textbf{MYO}          & \textbf{LV}                   & \textbf{RV}           & \textbf{MYO}          & \textbf{LV}           \\ \midrule
\multirow{4}{*}{ACDC}                       & UNETR                                              & 72.53 ± 0.81                  & 68.99 ± 0.42          & 83.11 ± 0.32                  & 19.02 ± 0.77          & 7.87 ± 0.09           & 8.55 ± 0.35           \\
                                            & +Synth ED/ES                                    & \textbf{77.73 ± 0.56}         & \textbf{76.43 ± 0.39} & \textbf{87.00 ± 0.38}         & \textbf{14.36 ± 1.25} & \textbf{6.44 ± 0.25}  & \textbf{6.82 ± 0.35}  \\ \cmidrule(l){2-8} 
                                            & Swin-UNETR                                         & 79.10 ± 0.26                  & 75.12 ± 0.28          & 86.96 ± 0.26                  & 12.70 ± 1.03          & 6.63 ± 0.21           & 7.21 ± 0.52           \\
                                            & +Synth ED/ES                                    & \textbf{82.70 ± 0.59}         & \textbf{81.27 ± 0.52} & \textbf{89.88 ± 0.38}         & \textbf{9.57 ± 0.25}  & \textbf{5.12 ± 0.29}  & \textbf{5.59 ± 0.56}  \\ \midrule
\multirow{4}{*}{M\&Ms}                      & UNETR                                              & 52.36 ± 4.52                  & 50.31 ± 6.30          & 64.23 ± 6.43                  & 37.06 ± 4.49          & 31.88 ± 4.26          & 35.20 ± 1.10          \\
                                            & +Synth ED/ES                                    & \textbf{66.55 ± 0.86}         & \textbf{65.99 ± 1.23} & \textbf{76.38 ± 1.62}         & \textbf{20.26 ± 0.73} & \textbf{16.06 ± 0.59} & \textbf{18.92 ± 0.68} \\ \cmidrule(l){2-8} 
                                            & Swin-UNETR                                         & 56.62 ± 1.33                  & 60.70 ± 0.71          & 72.47 ± 0.93                  & 50.33 ± 2.74          & 30.22 ± 2.04          & 35.40 ± 1.83          \\
                                            & +Synth ED/ES                                    & \textbf{73.73 ± 0.57}         & \textbf{74.48 ± 1.26} & \textbf{82.42 ± 0.25}         & \textbf{15.95 ± 1.96} & \textbf{10.56 ± 0.12} & \textbf{13.94 ± 0.98} \\ \midrule
\multicolumn{1}{l}{\multirow{4}{*}{M\&Ms2}} & UNETR                                              & 52.17 ± 2.62                  & 55.16 ± 2.71          & 71.71 ± 2.81                  & 31.55 ± 3.41          & 20.38 ± 2.23          & 22.80 ± 1.14          \\
\multicolumn{1}{l}{}                        & +Synth ED/ES                                    & \textbf{65.98 ± 0.93}         & \textbf{65.69 ± 0.42} & \textbf{80.16 ± 0.47}         & \textbf{21.14 ± 1.07} & \textbf{13.06 ± 0.22} & \textbf{14.66 ± 0.83} \\ \cmidrule(l){2-8} 
\multicolumn{1}{l}{}                        & Swin-UNETR                                         & 52.75 ± 2.26                  & 59.12 ± 0.52          & 76.33 ± 0.84                  & 41.07 ± 2.83          & 19.93 ± 0.87          & 21.78 ± 1.26          \\
\multicolumn{1}{l}{}                        & +Synth ED/ES                                    & \textbf{62.66 ± 2.04}         & \textbf{68.58 ± 0.70} & \textbf{81.96 ± 0.30}         & \textbf{22.90 ± 0.99} & \textbf{11.52 ± 0.25} & \textbf{14.41 ± 1.00} \\ \bottomrule
\end{tabular}%
}
\end{table}

Figure \ref{fig:corr} evaluates the ability of the proposed framework to generate specific cardiac volumes based on clinical priors. Pearson correlation coefficients vary across pathologies and guidance scales, ranging from moderate to strong ($r \in [0.66,0.94]$), with peak performance reaching $r=0.94$. However, higher CFG values also introduce a trade-off by potentially increasing the deviation from the absolute target volume while maintaining the desired volumetric trend.



\subsubsection{Segmentation evaluation.}
Table \ref{tab:nnunet_seg} summarizes the results using nnU-Net as segmentation backbone. On the internal ACDC dataset, 
the addition of synthetic data improves the LV DSC and significantly reduces HD across all structures. On the M\&Ms dataset, adding synthetic data improves the LV DSC by $3.40\%$ (from $84.87$ to $88.27$) and greatly reduces the LV HD by $6.12$~mm (from $14.57$ to $8.45$). Similar gains are observed on M\&Ms2, with a $4.73$~mm reduction in LV boundary error. These results suggest that our motion LDM generates anatomically coherent dynamics that help reduce the domain gap between different scanner vendors.

We extended our evaluation to UNETR and Swin-UNETR (Table \ref{tab:altseg}) to show that these improvements are not framework-specific. Vision Transformers typically require large-scale data to learn inductive biases and struggle with small datasets like ACDC. Our synthetic augmentation acts as a regularizer, providing performance boosts. On the M\&Ms dataset, the UNETR model achieves a $12.15\%$ improvement in LV DSC (from $64.23$ to $76.38$) when trained with our synthetic data. Overall, these findings demonstrate that the proposed framework provides a promising solution for 4D data augmentation that can benefit data-hungry models under domain shift. 


\section{Conclusion}

In this study, we developped a novel generative framework for 4D medical image synthesis, demonstrated through the explicit disentanglement of static anatomy and temporal dynamics via residual latent motion diffusion. By leveraging a semi-supervised VAE and latent motion priors, our approach generates paired 4D intensity volumes and segmentation masks. The results demonstrate that this approach significantly enhances the robustness and cross-vendor generalization of downstream tasks such as segmentation on cine CMR application. 
Future work will focus on extending this framework to a broader range of 4D medical imaging modalities characterized by complex spatiotemporal dynamics and annotation scarcity. Furthermore, the utility of the synthesized 4D sequences will be evaluated for additional clinical downstream tasks to further establish the universal benefits of this generative approach.


\bibliographystyle{splncs04}
\bibliography{references}
\end{document}